\documentclass[sigconf]{acmart}
\usepackage{float}
\usepackage{booktabs}
\usepackage{amsmath}
\usepackage{amsthm}
\usepackage{graphicx}
\usepackage{balance} 
\usepackage{needspace} 
\usepackage{booktabs}
\usepackage{algorithm}
\usepackage[noend]{algorithmic}
\usepackage{diagbox}
\usepackage{tabularx}
\usepackage{verbatim}
\usepackage{balance}

\begin{document}
\fancyhead{}

\title{WEASEL 2.0 - A Random Dilated Dictionary Transform for Fast, Accurate and Memory Constrained Time Series Classification}

\author{Patrick Sch\"afer}
\affiliation{%
  \institution{Humboldt-Universit\"at zu Berlin}
  \country{Germany}
}
\email{patrick.schaefer@hu-berlin.de}

\author{Ulf Leser}
\affiliation{
    \institution{Humboldt-Universit\"at zu Berlin}
    \country{Germany}  
}
\email{leser@informatik.hu-berlin.de}

\begin{abstract}

A time series is a sequence of sequentially ordered real values in time. Time series classification (TSC) is the task of assigning a time series to one of a set of predefined classes, usually based on a model learned from examples. Dictionary-based methods for TSC rely on counting the frequency of certain patterns in time series and are important components of the currently most accurate TSC ensembles. One of the early dictionary-based methods was WEASEL, which at its time achieved SotA results while also being very fast. However, it is outperformed both in terms of speed and accuracy by other methods. Furthermore, its design leads to an unpredictably large memory footprint, making it inapplicable for many applications. 

In this paper, we present WEASEL 2.0, a complete overhaul of WEASEL based on two recent advancements in TSC: Dilation and ensembling of randomized hyper-parameter settings. These two techniques allow WEASEL 2.0 to work with a fixed-size memory footprint while at the same time improving accuracy. Compared to 15 other SotA methods on the UCR benchmark set, WEASEL 2.0 is significantly more accurate than other dictionary methods and not significantly worse than the currently best methods. Actually, it achieves the highest median accuracy over all data sets, and it performs best in 5 out of 12 problem classes. We thus believe that WEASEL 2.0 is a viable alternative for current TSC and also a potentially interesting input for future ensembles.

\end{abstract}

\sloppy
\maketitle

\section{Introduction}

A time series (TS) is a collection of values sequentially ordered in time. TS emerge in many scientific and commercial applications, like weather observations, wind energy forecasting, industry automation, mobility tracking, etc. 
Research in TS is diverse and covers topics like storage, compression, clustering, etc.; see~\cite{esling2012time} for a survey. In this work, we study the problem of time series classification (TSC): Given a concrete TS, the task is to determine to which of a set of predefined classes this TS belongs to, the classes typically being characterized by a set of training examples. TSC has applications in many domains; for instance, it is applied to determine the species of a flying insect based on the acoustic profile generated from its wing-beat~\cite{PotamitisSchaefer2014}, or for identifying the most popular TV shows from smart meter data~\cite{greveler2012multimedia}. 

\begin{figure}[t]
\begin{centering}
\includegraphics[width=1\columnwidth]{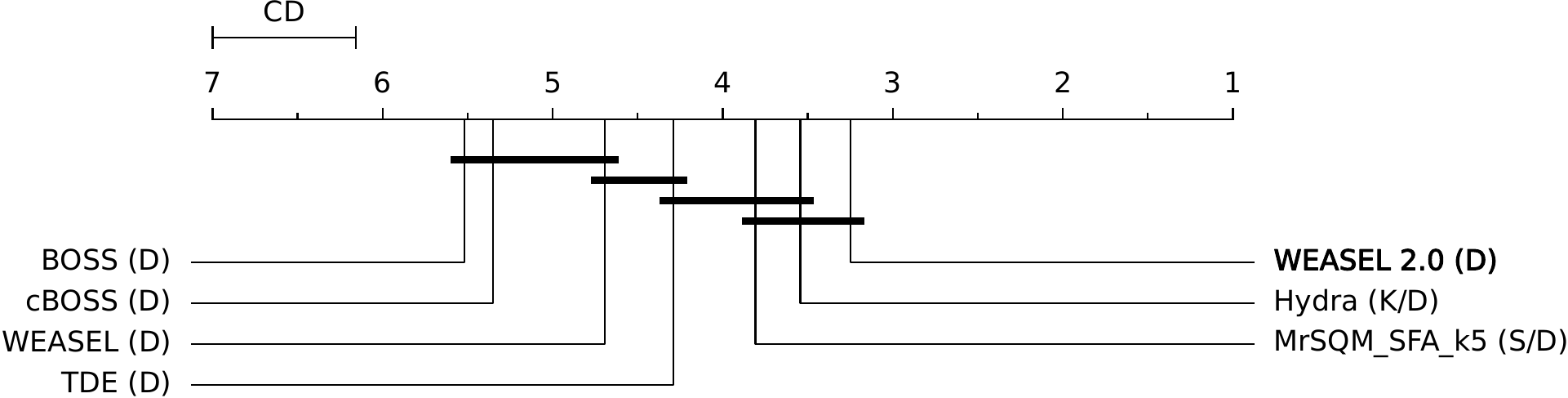}
\end{centering}
\caption{Critical difference plot on test accuracy for dictionary classifiers on 114 UCR datasets. WEASEL 2.0 is the most accurate dictionary classifier.  See Figure~\ref{fig:UCR_critical_accuracy} for a comparison to all SotA classifiers.\label{fig:UCR_critical_accuracy_subset}}
\end{figure}

To date, among the most accurate approaches are kernel-based methods~\cite{dempster2020rocket, dempster2021minirocket, tan2022multirocket}, shapelets~\cite{guillaume2022random} and hybrids~\cite{middlehurst2021hive, shifaz2020ts} (ensembles of base TSC). 

Dictionary-based methods, such as BOSS, WEASEL, TDE or MUSE~\cite{schafer2014boss, middlehurst2021temporal, schafer2017multivariate, schafer2017fast} have historically been among the most accurate classifiers~\cite{bagnall2016great,ruiz2021great}. 
Conceptually, dictionary approaches extract phase-independent subsequences by sliding a window over TS. Each window is
transformed into a word, and the frequency of repeating patterns is recorded. A classifier can be learned on the resulting feature vector. WEASEL~\cite{schafer2017fast}, which is the basis of this work, uses SFA~\cite{SchaferH12} for word generation, and a Ridge regression classifier. The symbolic transformation SFA~\cite{SchaferH12} is used in all SotA dictionary-based classifiers. It applies a Fourier transform to subsequences, and discretizes frequencies into symbols.  
Yet recently, these dictionary methods have fallen behind in accuracy, and their major drawback is their potentially huge memory-footprint, due to a large variance in generated words. 

A recent theme in TSC is to build large, but size-controlled features space of several thousand features, and use these as input to train a linear RIDGE regression classifier. The large feature space is generated from a single high bias transform, by building an ensemble on randomized sets of hyper-parameters~\cite{dempster2020rocket, dempster2021minirocket, tan2022multirocket, guillaume2022random}. Both in combination generate a low bias and low variance transformation usable by a linear classifier, for high accuracy and fast prediction. Another recent discovery is the use of a dilation operation applied to a filter, such as the convolutional filter (ROCKET~\cite{dempster2020rocket, dempster2021minirocket, tan2022multirocket, dempster2022hydra}) or a sliding window operation (R-DST~\cite{guillaume2022random}). Dilation adds a gap between values, effectively increasing the size of the receptive field of the filter, and also operating similar to a down-sampling operation. Thus, dilation offers features at multiple scales.

In this work, we present a complete overhaul of the dictionary transform WEASEL, using two state-of-the-art techniques: (a) randomized hyper-parameter ensemble and (b) dilation. WEASEL 2.0 addresses many of the shortcomings of current dictionary-based transforms, including the memory foot-print and it inferior accuracy. 

Figure~\ref{fig:UCR_critical_accuracy_subset} shows the advances made by WEASEL 2.0 in comparison to SotA dictionary transforms (Hydra~\cite{dempster2022hydra} is a hybrid between kernel-based and dictionary, and MrSQM~\cite{mrsqm2022} is a hybrid between Shapelet and Dictionary). On $114$ UCR datasets it is the best of its class, and significantly more accurate than its predecessor WEASEL, and furthermore it has a predictable memory footprint (see Figure~\ref{fig:UCR_memory}). Its dictionary is of only some tens of thousand of features, which makes it similar in size to kernel-based methods.

In summary, our contributions are as follows:
\begin{enumerate}
    \item We introduce a novel dilation mapping. This allows to turn any subsequence-based method into a dilated algorithm, i.e. it is not limited to dictionary-based methods. This mapping can be implemented using just two lines of python code, and can be applied prior to the down-stream classification task. 
    \item We introduce changes to the symbolic transformation SFA~\cite{SchaferH12}, used in word generation, to significantly reduce the feature space to just $256$ words, including a novel variance-based feature selection strategy. SFA is used in the SotA classifiers and ensembles such as~\cite{mrsqm2022, middlehurst2021temporal, middlehurst2021hive, shifaz2020ts}, thus any improve in SFA potentially benefits these, too.
    \item The refined word generation in combination with random ensembling over multiple hyper-parameters produces a predictable size of the feature space. Thereby, we solve the major shortcoming of dictionary methods.
    \item Through extensive experiments on the UCR datasets, we show that WEASEL 2.0 is as fast and not significantly different in accuracy than the non-ensemble SotA, namely ROCKET~\cite{dempster2020rocket}, MiniRocket~\cite{dempster2021minirocket}, MultiRocket~\cite{tan2022multirocket}, or R-DST~\cite{guillaume2022random}. Furthermore, WEASEL 2.0 is the best dictionary-based method.
\end{enumerate}

The rest of the paper is organized as follows:
In Section~\ref{sec:background} we present the background and in Section~\ref{sec:related_work} the related work on time series, classification, and dictionary approaches. Section~\ref{sec:weasel2.0} presents the novel WEASEL 2.0. Section~\ref{sec:experiments} shows our experimental evaluation and Section~\ref{sec:conclusion}  concludes the paper.

\section{Background}~\label{sec:background}

We will first formally define the basic concepts. 
\begin{definition}
\emph{Time Series (TS)}: A time series $T = \left(t_1, t_2,\dots, t_n \right)$ is an ordered sequence of $n$ real values. We denote the $i$-th value of $T$ by $t_{i}$.
\end{definition}

Such a TS is also refereed to as \emph{univariate time series}. If each point represents multiple variables (e.g. humidity, temperature and pressure) we call it a \emph{multivariate data series}.

Most SotA TSC algorithms make use of subsequences of the data. 
\begin{definition}
\emph{Subsequence}: A subsequence $T_{i,l}$ of $T = (t_1, \dots, t_n)$, with $1 \leq i \leq n$ and $1 \leq i+l \leq n$, is a subseries of length $l$, consisting of $l$ contiguous points from $T$ starting at offset i: $T_{i,l} =(t_i,t_{i+1},...,t_{i+l-1})$
\end{definition}
We may extract subsequences from a TS by the use of a sliding window. 
\begin{definition}
\emph{Sliding Window}: 
A time series $T$ of length $n$ has $(n-l+1)$ sliding windows of length $l$, when increment is $1$, given by: $$sliding\_windows(T)=\{T_{1,l},\dots,T_{(n-l+1),l}\}$$
\end{definition}

\emph{Time series classification (TSC)} is the task of predicting a class label for a TS whose label is unknown. A TS classifier is a function that is learned from a training dataset of discrete, labeled time series, takes an unlabeled time series as input and outputs a label. 

\begin{definition}
\emph{Dataset}: A dataset $D=\left(T^{(i)}, y^{(i)}\right)_{i \in [1\dots m]}$ is a collection of $m$ time series, each assigned to one of a predefined set of classes $Y$.  We denote the size of $D$ by $m$, and the $i^{th}$ instance by $T^{(i)} \in \mathbb{R}^{n}$, and its label by $y^{(i)} \in Y$.
\end{definition}

A common operation to reduce the length of a very large TS, is to take every $d$-th value, dropping all other values, and referred to as down-sampling.

\begin{definition}
\emph{Down-Sampling}: 
A time series $T$ of length $n$ is down-sampled by factor $d$, by taking every $d$-th value from $T$: $$down\_sample(T, d) = T_{1::d} = (t_1,t_{1+d}, t_{1+2 \times d}, \dots)$$
\end{definition}

\begin{definition}
\emph{Dilated Subsequence}: A dilated subsequence of $T$ with dilation $d$ and offset $i$, with $1 \leq i \leq n$ and $1 \leq i+l \times d \leq n$, is a subseries of length $l$, obtained from down-sampling the time series starting from offset $i$ and then extracting the first $l$ continuous points: 
$$dilated\_window(T,i, d) = \left(t_i, t_{i+d\times 1}, t_{i+d\times 2},...,t_{i+d \times (l-1)}\right)$$
\end{definition}



\section{Related Work}\label{sec:related_work}

In this section, we will first introduce the techniques used in time series classification (Section~\ref{sec:tsc}). Next, we will put the focus on dictionary-based methods, the symbolic transformation SFA (Section~\ref{sec:dictionary}), and WEASEL (Section~\ref{sec:TSCWEASEL}).

\subsection{Time Series Classification (TSC)}\label{sec:tsc}

The techniques used for TSC can be broadly categorized into the following categories~\cite{bagnall2016great}. We present a brief overview of algorithms, based on first movers and state-of-the-art in their field.

\begin{enumerate}
    \item \emph{Distance based} classifiers  use a distance function, to measure the similarity, and a classification algorithm on the distances. Historically, distance functions have been mostly used with nearest neighbor (NN) classifiers. Commonly, 1-NN Dynamic Time Warping (DTW)~\cite{rakthanmanon2012searching} was used as a baseline in comparisons~\cite{lines2014time,bagnall2016great}, but is now significantly worse than SotA~\cite{bagnall2016great, ruiz2021great}.
    
    Typically, these techniques work well for short but fail for noisy or long TS~\cite{schafer2014boss}. Furthermore, DTW has a computational complexity of $O(n^{2})$ for TS of length $n$. Recent advances include the Matrix Profile Distance (MPDist)~\cite{gharghabi2018matrix}, which defines two time series to be similar, if they share many phase-independent subsequences. 

    \item \emph{Exploratory transformations} are a popular recent theme. These extract descriptive statistics as features from time series (subsequences) to be used in classifiers. Several tool-kits exist for extracting features, including hctsa~\cite{christ2018time}, with over 7700 features. catch22~\cite{lubba2019catch22} contains a subset of $22$ dominant features from hctsa. \emph{tsfresh}~\cite{christ2018time} is a collection of roughly $800$ features.

    \item \emph{Shapelets} are phase independent discriminatory subsequences, which \emph{presence} or \emph{absence} can be an indicator of a class of a TS. The expression of a shapelet is found by sliding the shapelet across the TS, and minimizing the Euclidean distance between each subsequence in the TS and the shapelet. The currently most accurate Shapelet approach is R-DST~\cite{guillaume2022random}. It combines randomization on hyper-parameters with dilation to increase its diversity. A total of $10k$ shapelets with three features each are extracted, and are fed into a RIDGE regression model.     
    MrSEQL~\cite{agarwal2021ranking} is an ensemble classifiers that looks for the absence of presence of patterns. But other than the previous method, which minimizes the distance between raw subsequences, MrSEQL discretizes subsequences into words using SFA~\cite{SchaferH12} and searches for matches of words. A set of discriminative words is selected through Sequence Learner (SEQL)~\cite{ifrim2011bounded}. 

    \item \emph{Dictionary approaches} use phase-independent subsequences by sliding a window over time series, too. But rather than to measure the distance to a subsequence, as in shapelets, each window is transformed into a word, and the frequency of occurrence of repeating patterns is recorded. These methods discriminate based on the frequency of patterns. BOSS~\cite{schafer2014boss} converts each sliding window into a word by the use of the Symbolic Fourier transformation (SFA)~\cite{SchaferH12}. A classifier can be built using a non-symmetric distance function in combination with a 1-NN classifier. Temporal Dictionary Ensemble (TDE)~\cite{middlehurst2021temporal}
    is an ensemble of 1-NN classifiers, each transforming the time series into a histogram of word counts using SFA. Conceptually, TDE combines properties of different flavors of dictionary classifiers, like WEASEL~\cite{schafer2017fast}, or SpatialBOSS~\cite{large2019time}.    
    
    \item \emph{Kernel/Convolution} classifiers extend on the idea of Shapelets. Shapelets can be realised through a convolution operation, followed by a min-pooling operation on the array of windowed Euclidean distances. This was first observed in~\cite{grabocka2014learning}. The main difference between convolutions and shapelets is that shapelets are searched for from the candidate space of subsequences extracted from the training data. Yet, convolution filters are found from the entire space of possible real-values. The most well known approach is ROCKET~\cite{dempster2020rocket}, with its successors MiniROCKET~\cite{dempster2021minirocket}, MultiROCKET~\cite{tan2022multirocket}, and Hydra~\cite{dempster2022hydra}. ROCKET generates tens of thousands of randomly parameterized convolutional kernels, and applies two pooling operations to the output: the maximum and the proportion of positive values. These are used as features in RIDGE regression. The first extension MiniROCKET is significantly faster with an equal accuracy, and reduces the randomization of parameters of ROCKET, making it almost deterministic. MultiROCKET builds on MiniROCKET adding new pooling operations and first order differences. MultiROCKET is one of the most accurate classifiers to date~\cite{tan2022multirocket}.
    
    \item \emph{Hybrid:} The most accurate current TSC algorithms are ensembles of multiple types of classifiers. HIVE-COTE v1~\cite{lines2016hive} incorporates classifiers from five domains. It was recently updated to version 2.0~\cite{middlehurst2021hive} to address scalability issues, and reflect recent innovations. It refined the ensemble to use more accurate classifiers, such as ROCKET (Arsenal~\cite{middlehurst2021hive}) or TDE. To date, this is the most accurate classifier. But it is one to two orders of magnitude slower than kernel-based classifiers while not being significantly more accurate~\cite{tan2022multirocket}.
    
    \item \emph{Deep Learning:} InceptionTime~\cite{ismail2020inceptiontime} is an ensemble of deep convolutional neural networks based on the Inception architecture. According to its authors, it is currently the most accurate deep learning approach for TSC. LSTM-FCN~\cite{karim2017lstm} combines a recurrent neural network with a fully connected neural network. At the time of its publication, it was among the most accurate approaches.
\end{enumerate}

\subsection{Dictionary-based Approaches}\label{sec:dictionary}

\begin{figure}
\begin{centering}
\includegraphics[width=1\columnwidth]{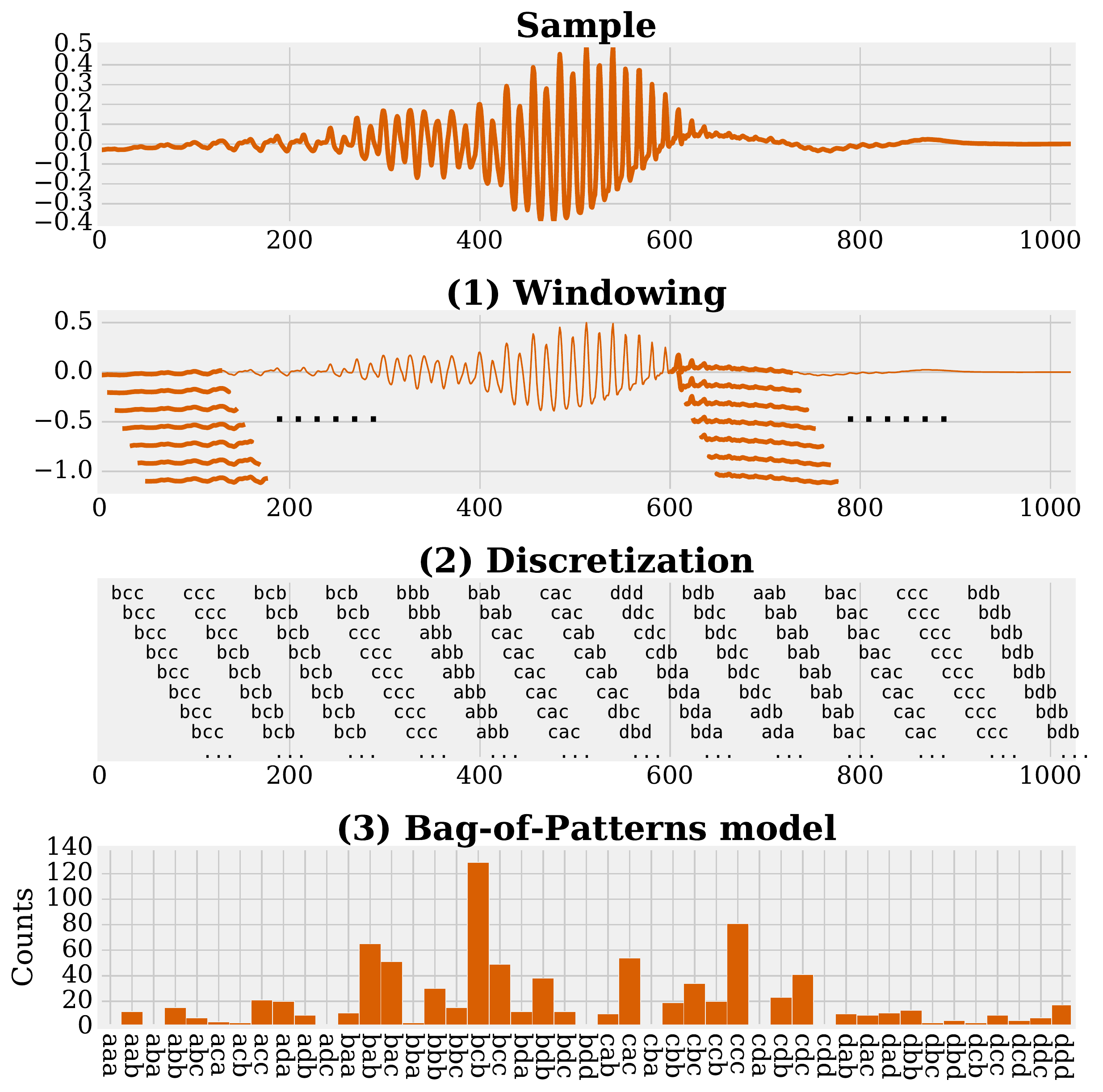}
\par\end{centering}
\caption{Transformation of a TS into the dictionary-based model (from ~\cite{schafer2017fast}) using overlapping windows (second to top), discretization of windows to words (second from bottom), and word counts (bottom).\label{fig:transformation}}
\end{figure}

Algorithms following the dictionary-based model build a classification function by:
\begin{enumerate}
    \item Extracting subsequences, aka \emph{windows}, from a TS;
    \item Transforming each window of real values into a discrete-valued \emph{word} (a sequence of symbols over a fixed alphabet); 
    \item Building a sparse feature vector from word counts, and 
    \item Finally using a classification method from the machine learning repertoire on these feature vectors. 
\end{enumerate}
Figure~\ref{fig:transformation} illustrates these steps from a raw time series to a dictionary model using overlapping windows.

Dictionary-based methods differ in the concrete way of transforming a window of real-valued measurements into discrete words (discretization). For example, the basis of the BOSS model, TDE~\cite{middlehurst2021temporal}, or MrSQM~\cite{mrsqm2022}, is a symbolic representation called SFA. SFA works as follows~\cite{SchaferH12} : 

\begin{enumerate}
    \item Values in each window are normalized to have standard deviation of $1$ to obtain amplitude invariance. 
    \item Each normalized window of length $w$ is subjected to dimensionality reduction by the use of the truncated Fourier transform, keeping only the first $l<w$ coefficients for further analysis. This step acts as a low pass filter, as higher order Fourier coefficients typically represent rapid changes like dropouts or noise. 
    \item Each coefficient is discretized to a symbol of an alphabet of fixed size $\alpha$ to achieve further robustness against noise. 
\end{enumerate}

\begin{figure}
\begin{centering}
\includegraphics[width=1\columnwidth]{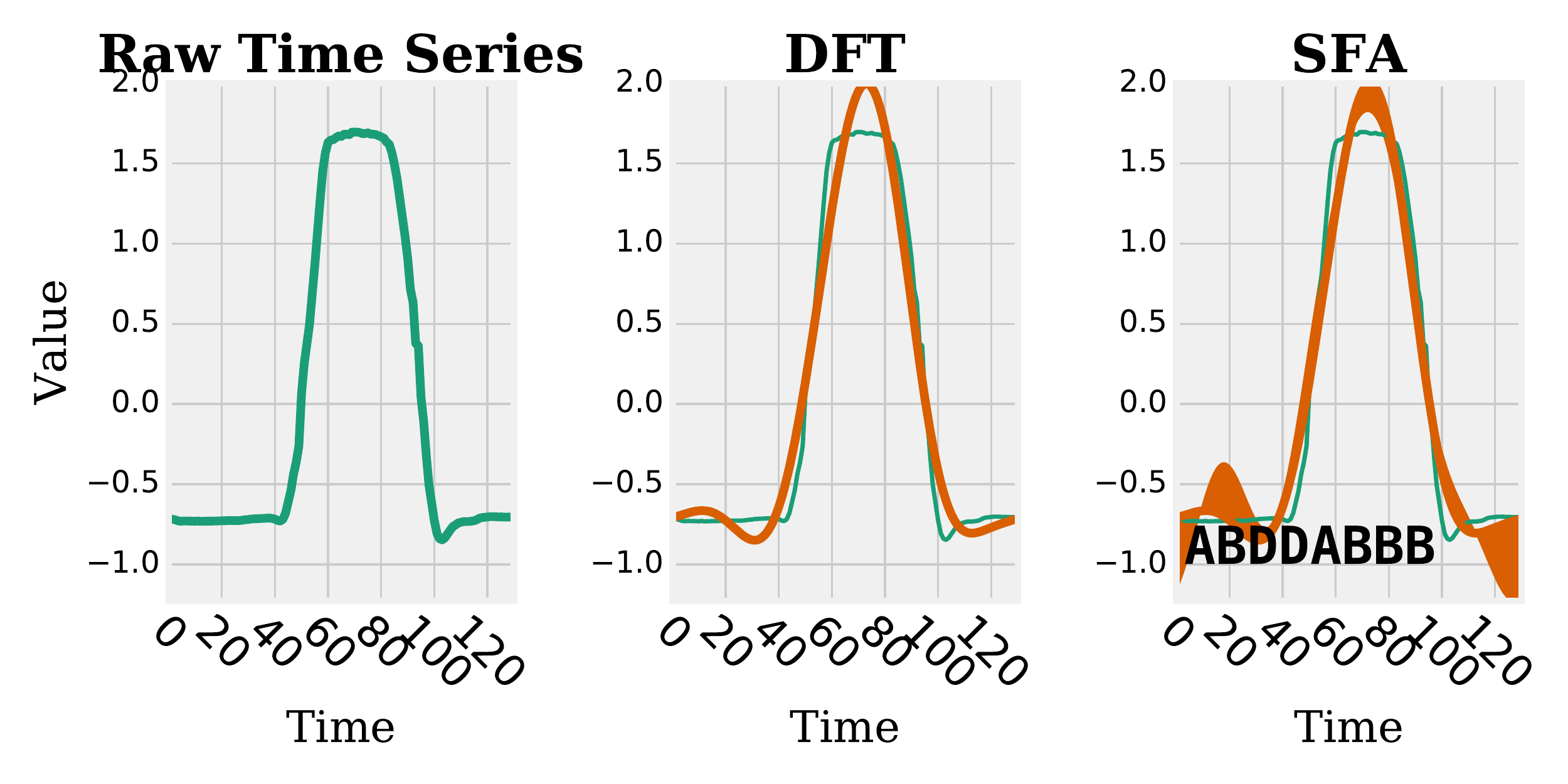}
\par\end{centering}
\caption{The Symbolic Fourier Approximation (SFA) (from ~\cite{schafer2017fast}): A time series~(left) is approximated using the truncated Fourier transform~(center) and discretized to the word \emph{ABDDABBB}~(right) with the four-letter alphabet ('a' to 'd'). The inverse transform is depicted by an orange area (right), representing the tolerance for all signals that will be mapped to the same word.\label{fig:SFATransform}}
\end{figure}

Figure~\ref{fig:SFATransform} exemplifies this process from a window of length $128$ to its DFT representation, and finally the word \emph{ABDDABBB}.

\subsection{WEASEL 1.0}\label{sec:TSCWEASEL}

WEASEL, as published in~\cite{schafer2017fast}, refined the dictionary approaches to add supervision using class labels. It is the basis of the WEASEL 2.0 model presented in this paper. Thus, we will refer to WEASEL as WEASEL 1.0 in the following.

\begin{figure}
\begin{centering}
\includegraphics[width=1\columnwidth]{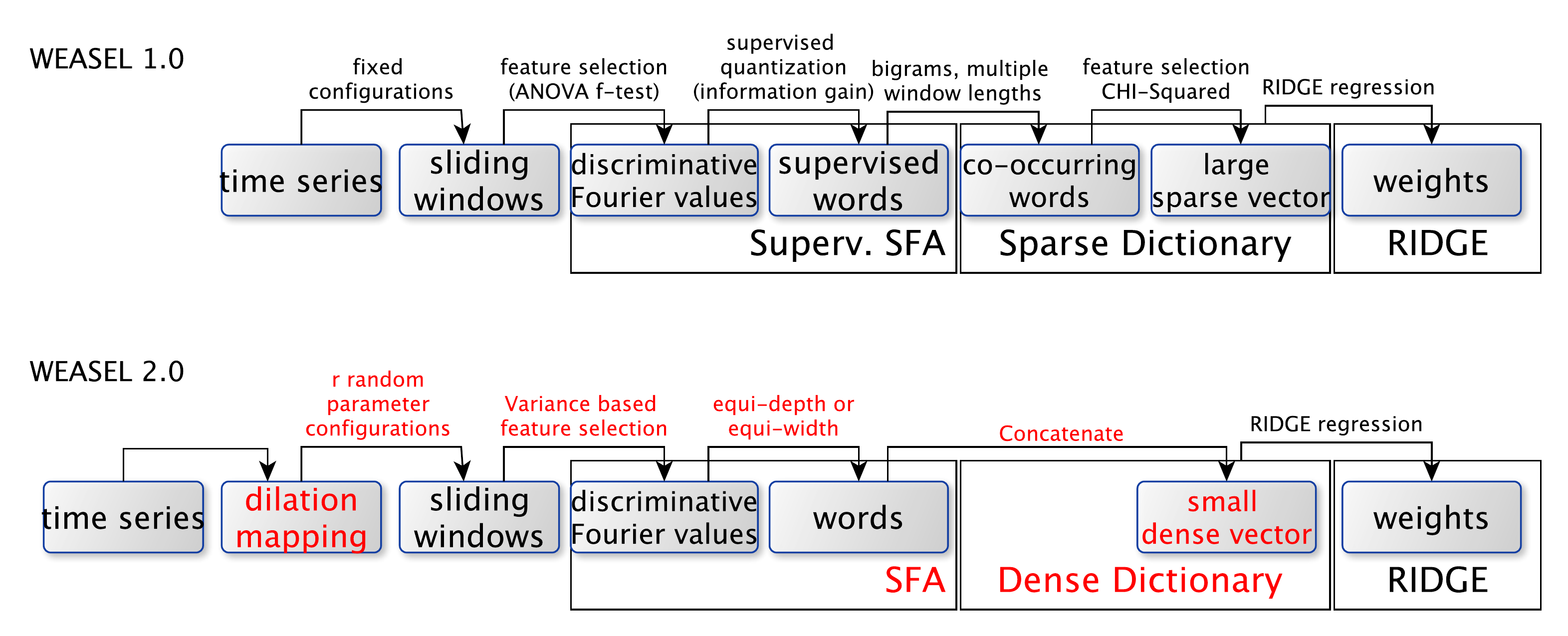}
\par\end{centering}
\caption{The WEASEL 1.0 (from~\cite{schafer2017fast}) and WEASEL 2.0 pipelines. Differences are highlighted in red. WEASEL 2.0 adds a dilation mapping prior to the sliding window operator, carefully chooses parameters to control the size and memory consumption of the dictionary, and adds randomization to increase variance.\label{fig:WEASEL-Pipeline}}
\end{figure}

WEASEL 1.0 is composed of the building blocks depicted in Figure~\ref{fig:WEASEL-Pipeline}: a supervised SFA representation for discriminative word generation, and a large sparse dictionary of word counts. First, WEASEL extracts normalized subsequences (windows) of varying lengths from a TS. Next, each window is approximated using the Fourier transform, and the real and imaginary Fourier values are kept that best separate TS from different classes, determined using the ANOVA F-test. These selected Fourier values are then discretized into a word based on information gain binning, using class labels to choose those discretization boundaries to best separate the TS classes; This process is similar to a decision tree split. Finally, a single, large sparse dictionary patterns is built from the words (unigrams), neighboring words (bigrams), over all chosen window lengths. To filter irrelevant words, and reduce the size of the dictionary, a Chi-squared test is applied. A RIDGE regression classifier is trained on the retained dictionaries.

\begin{figure*}[ht]
\begin{centering}
\includegraphics[width=2.0\columnwidth]{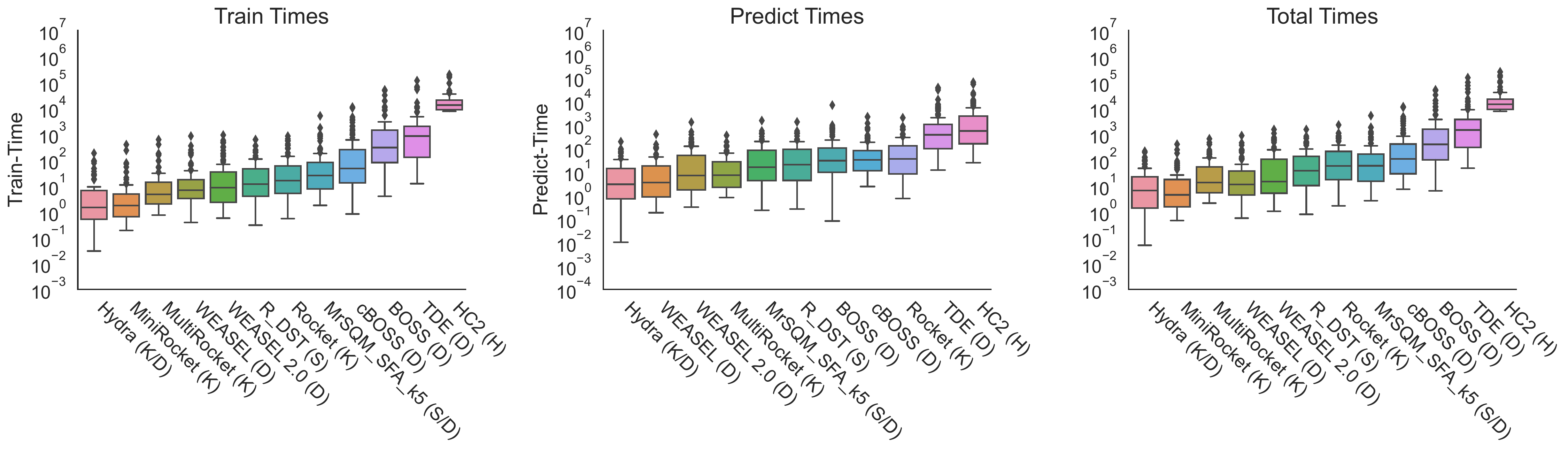}
\end{centering}
\caption{Runtime (Training, Prediction, Total) on the 114 UCR datasets by method in seconds.\label{fig:UCR_runtime}}
\end{figure*}

WEASEL 1.0 is still among the fastest classifiers, see Figure~\ref{fig:UCR_runtime}, but its accuracy is significantly worse than the SotA, and its dictionary can result in excessive memory consumption (Figure~\ref{fig:UCR_memory}).

\section{WEASEL 2.0 - A Random Dilated Dictionary Classifier}~\label{sec:weasel2.0}

We observed two design issues (DI) present in current dictionary classifiers, not just limited to WEASEL 1.0 but also MrSQM, TDE, BOSS:
\begin{enumerate}
    \item \emph{DI 1: Memory footprint}: A major shortcoming of dictionary based approaches is that the feature space is \emph{huge, but sparse}. This allows for high accuracy classification using linear classifiers, but results in large amounts of memory to be allocated even for small datasets. 
    \item \emph{DI 2: Sensitivity}: A negative effect of supervision in word generation is that a minor change in two very similar subsequences can result in two distinct words, when it causes a Fourier value close to a discretization boundary to change. This is typically compensated by increasing the number of words generated using different parameterizations of SFA. If we however restrict the dictionary size, the increased sensitivity to small changes can deteriorate accuracy.
\end{enumerate}

Our main goals in the design of WEASEL 2.0 were thus to constrain the memory consumption, and add robustness (improve the sensitivity) in word generation. WEASEL 2.0 differs from WEASEL 1.0 in multiple aspects highlighted in red in Figure~\ref{fig:WEASEL-Pipeline}:

\begin{enumerate}    
\item Dilation is applied to each time series using a dilation mapping (see Section~\ref{sec:dilation}). This adds state of the art techniques to WEASEL;
\item A fixed-size dictionary of just $256$ words is generated for each parameter-set to reduce the impact of minor changes in extracted windows (see Section~\ref{sec:dictionary_construction}). This addresses DI 1;
\item A variance-based real and imaginary Fourier value selection strategy for SFA is introduced (see Section~\ref{sec:word_generation}). We add first order differences of the time series, just as in MUSE~\cite{schafer2017multivariate}. This addresses DI 2;
\item We apply randomization to choose parameter configurations and thereby increase variance (see Section~\ref{sec:ensemble}). This addresses DI 2;
\item The final fixed-size feature vector is small and used for classification through a fast RIDGE regression classifier (see Section~\ref{sec:pseudocode}).
\end{enumerate}

\subsection{Dilation Mapping}~\label{sec:dilation}

Dilation is a technique that allows a filter, such as a convolution filter or sliding window, to cover more of the time series data by inserting \emph{gaps} between the entries in the filter. These gaps allow the filter to increase the size of the receptive field, while keeping the total number of values constant. For example, a dilation of $d=2$ would insert a gap of $1$ between every pair of values. This effectively doubles the size of the receptive field, and allows the filter to process the data at different scales, similar to a down-sampling operation.

Dilation is one core mechanism related to the recent increase in scalable and accurate classifiers. In TSC it was first proposed by the ROCKET-family of classifiers. Dilation provides analysis at different scales, while keeping the number of values of the filter constant. In ROCKET filter sizes of $7, 9$ or $11$ are common. In prior methods, the window size was increased up to hundredths of values, increasing computations and memory footprint, and thereby reducing scalability. 

Introducing dilation to an algorithm involves a (major) rewrite of its code-base. Large TSC libraries such as sktime~\cite{loning2019sktime}, contain dozens of classifier. Touching every classifier, is not feasible.
One of our main contributions is to show that a simple transformation on the time series - which effectively reorders all values - will result in a dilation operation to be applied by the down-stream classification model. This transformation can be applied as a pre-processing step, and can be achieved using just \emph{two lines of python code}. Using this transformation, we do not have to modify the code of any sliding window-based model (such as shapelets, dictionaries, interval) apart from adapting its hyper-parameters.

\begin{figure}
\begin{centering}
\includegraphics[width=1\columnwidth]{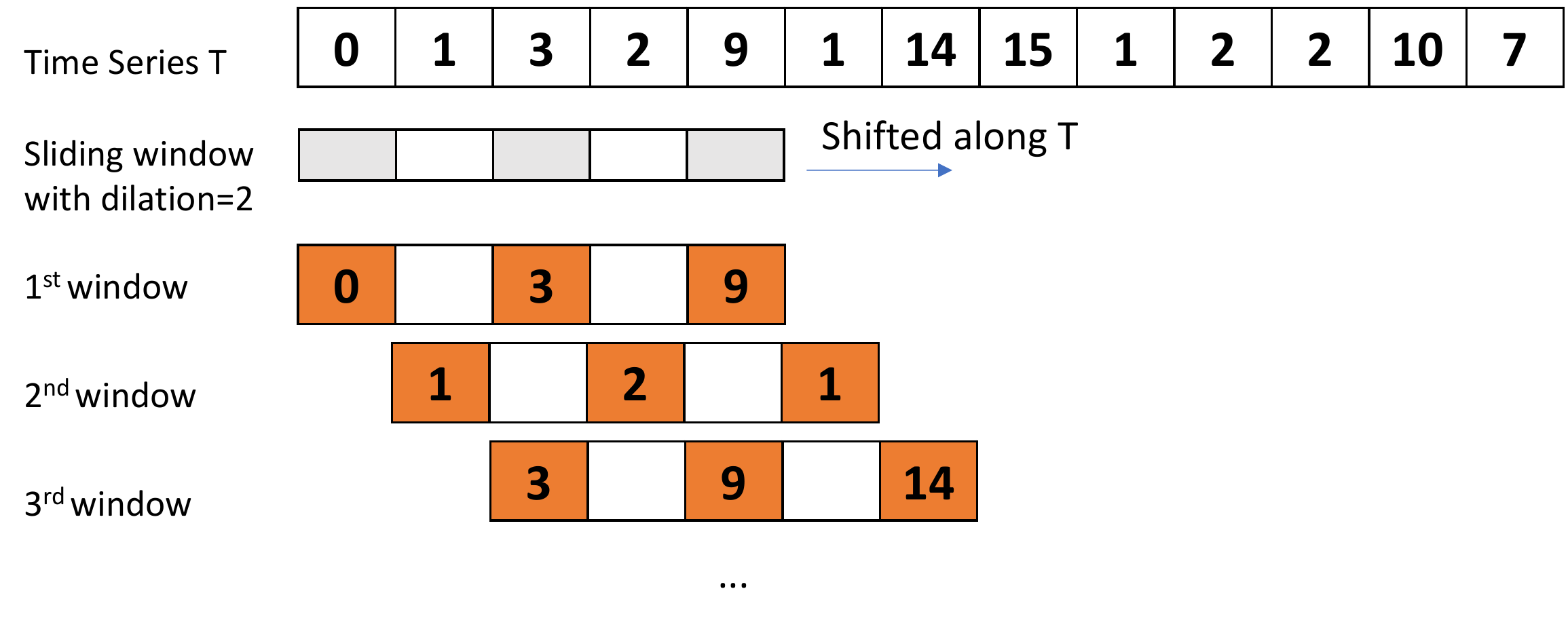}
\par\end{centering}
\caption{A dilated windowing operation applied to a time series with $d=2$.\label{fig:dilation}}
\end{figure}

Figure~\ref{fig:dilation} illustrates the concept of dilation, when used in conjunction with sliding windows. A dilation of $d=2$ is added to the sliding window, effectively adding a gap of $1$ between each value. The dilated sliding window then filters every $d^{th}$ value of the time series, and finally the dilated window can be fed into the downstream processing task. 

\begin{figure*}
\begin{centering}
\includegraphics[width=2.\columnwidth]{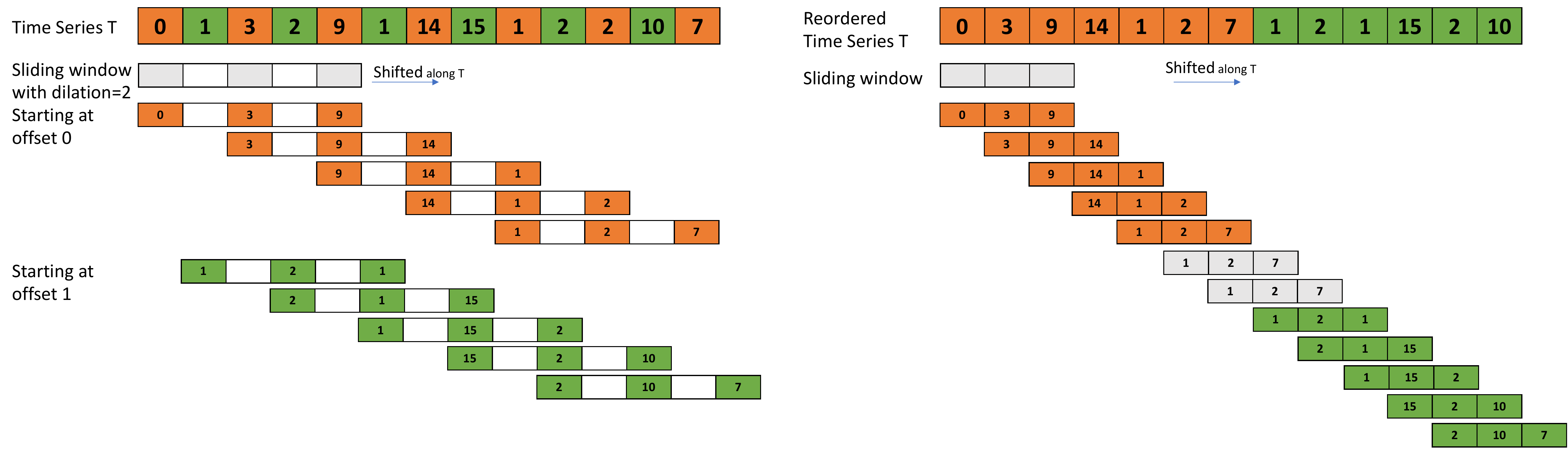}
\par\end{centering}
\caption{(Left): A dilated sliding window operation applied to an input time series. This yields two sets of subsequences starting at uneven (orange) and even (green) offsets. (Right): Reordering the green and orange values leads to a sliding window operation with equivalent results as when using dilation.\label{fig:reordering}}
\end{figure*}

When coloring the windows extracted from a dilated sliding window with dilation $d=2$, we find two sets of overlapping subsequences (Figure~\ref{fig:reordering}). The first set of windows starts at every even offsets (orange). The second set starts at every even offset (green). Within the even or uneven windows, each window overlaps with its successor by all but the \emph{first} and \emph{last} value. This is conceptually equivalent to an ordinary sliding window on a re-ordered TS. 
In the case of $d = 2$, we first need to take every uneven index ($1, 3, 5, \dots $), and then concatenate this subsequence to every even index ($2, 4, 5, \dots $). 
The same concept applies to higher dilation factors $d>2$: we build groups for the first $d$ indices, take every $d$-th value in each, and concatenate the resulting $d$ subsequences to form a reordered new TS.

\begin{theorem}
    The windows extracted by a dilated sliding window, with dilation $d$, are a subset of those windows extracted from a reordered time series $T'$, where $T'$ is constructed by a $d$-rate down-sampling of indices $i \in [0,..,d)$:
    $$\textit{dilated\_window}(T, d) \subset \textit{sliding\_window}\left(T_{0::d} \cup T_{1::d} \cup T_{(d-1)::d}\right)$$
\end{theorem}

\begin{algorithm}[htb]
    \caption{apply\_dilation (Time Series $T$ and dilation rate $d$)}
    \label{alg:dilation}
    \begin{algorithmic}[1]            
        \FOR {$i$ in $range(0, d)$}
            \STATE $T' = \textit{np.concatenate}([T', T[i::d]])$
        \ENDFOR
        \RETURN $T'$
    \end{algorithmic}
\end{algorithm}

We apply this dilation mapping in the implementation of WEASEL 2.0.

The main advantage of this operation is that we do not have to alter any aspect of the downstream algorithm. We simply reorder the input time series, and train the original model. In principle this operation allows to turn any algorithm into a dilated algorithm. The major disadvantage is that there are a few additional windows generated at the intersections of the re-ordering (the two gray subsequences on the right in Figure~\ref{fig:reordering}). The longer the time series becomes, however, these hardly have any impact. 

The dilation mapping is a linear time and space operation in the length of the time series. The dilated time series can be discarded once the classifier has been trained.

\subsection{Dictionary Construction}\label{sec:dictionary_construction}

Several methods achieve SotA results~\cite{dempster2020rocket, tan2022multirocket, dempster2021minirocket} using only some thousands of features for high accurate classification though linear models.
The feature space of WEASEL 1.0 can grow up to millions of features for even small TS. While statistical feature selection (Chi-squared) has been applied to counter-act this, the number of features above a p-value remains unpredictable. We address this, and build predictable size dictionaries through two design decisions: 

\begin{enumerate}
    \item \emph{Dense Dictionary}: We build a high bias, dense dictionary representation of the time series containing just $256$ words. Thus, all subsequences are mapped to one of these words, increasing robustness. However, the low number of words limits the overall classification accuracy.
    \item \emph{Ensembling Parameters}: We increase diversity (variance) and reduce bias, through ensembling over multiple parameter configurations ($50, 100$ or $150$) using randomization. This results in a controlled fixed-size dense dictionary of size $256 \cdot 50 \approx 12.8k$ to $256 \cdot 150 \approx 38k$,
    
\end{enumerate}

\paragraph{Dense Dictionary}

SFA generates its dictionary over an alphabet size $\alpha$ and a word length $l$.
The upper bound on the dictionary size is computed as $\alpha ^ l$.
All subsequences are mapped to one of these words. Using bigrams explodes this feature space to a theoretical upper bound of $\alpha ^ {2l}$. 

For the default parameters in WEASEL 1.0 ($\alpha=4$, $l=6$, $bigrams=True$) this feature space can create $4 ^ {12} = 16M$ distinct words. However, as each subsequence can only generate one word and one bigram, the resulting feature space is sparse. While this diversity can be favorable to identify single words to distinguish between time series, it required large amounts of memory and increases time for training or prediction.

To build a dense dictionary with guaranteed memory consumption, we decided to default $\alpha=2$, $l=8$ and disable bigrams. This reduces the size of the dictionary to just $2^8=256$. I.e. all subsequences are mapped to one of $256$ different words. Technically, this is implemented by using an array of $256$ Integers. However, this tiny feature space makes it hard for any linear classifier to find discriminating features, resulting in high bias. To reduce bias and increase variance, we ensemble different hyper-parameter configurations using randomization (Section~\ref{sec:ensemble}). In the next section we introduce a variance-based of Fourier values selection strategy to derive robust words using SFA.

\subsection{Robust Word Generation}\label{sec:word_generation}

The supervision introduced in SFA, such as \emph{information-gain} binning, \emph{ANOVA F-test} coefficient selection, and CHI-squared word selection, provide high variance, but are  sensitive to minor changes in values  of the time series. This would make SFA very brittle when combined with reducing the overall number of words generated to $256$. 

To compensate for this, we introduce a novel \emph{variance}-based Fourier value selection strategy. This strategy first computes the variance within each real and imaginary value of the Fourier transform, and then chooses the top $l$ real and imaginary values by highest variance. The larger the variance, the larger the discretization interval (bins) may become. Thus, we avoid minor value changes in similar subsequences leading to different words. 

Other than in WEASEL 1.0, which uses information-gain to derive bins, in WEASEL 2.0 we restrict SFA to only use \emph{equi-width} or \emph{equi-depth} binning, as these showed to be the most robust against overfitting.

\subsection{Ensemble Generation}\label{sec:ensemble}

Our goal was to restrain the number of distinct features (words) to a fixed, predictable number, with which a linear RIDGE classifier can still find distinctive features with high accuracy. We achieve this through randomization on hyper-parameters, where the number of drawn parameter-configurations, aka ensemble size in the following, determines the overall number of features generated per TS. WEASEL 2.0 has three key hyper-parameters to set:
\begin{enumerate}
    \item \textbf{Minimal window length} $w\_{min}$: Typically defaulted to $4$
    \item \textbf{Maximal window length} $w\_{max}$: Typically chosen from $24$, $44$ or $84$ depending on the time series length.
    \item \textbf{Ensemble size} $r_{max}$: Typically chosen from $50, 100, 150$, to derive a feature vector of roughly $20k$ up to $70k$ features (distinct words).
\end{enumerate}    

Other than ROCKET or R-DST, which limit the size of its filter to $\{7,9,11\}$, our window size is in a range of $4$ up to $84$. For a meaningful approximation via SFA, we require large windows. SFA is applied to the windows, extracts Fourier values, and generates words of length $8$ and $\alpha=2$. If we used the Rocket size of filters, the SFA transformation would eventually be a 1:1 mapping from values to symbols, providing no benefits from the Fourier transform.

We use a simple rule of thumb as default for $r_{max}$ based on the dataset size $m$ and ts length $n$:
$$
    r_{max} = \left\{
        \begin{array}{lr}
        50,  & \text{if } m \leq 250\\
        100, & \text{if } (m > 250) ~\&~ (n \leq 100)\\
        150, & \text{else } 
        \end{array}\right.
$$
The rationale is to use a larger ensemble, the more or the longer the time series in the dataset become. 

We then randomly initializes one parameter-set for each of the $r_{max}$-many configurations using randomization:
\begin{enumerate}
    \item \textbf{Window length} $w$: Randomly chosen from interval $[w\_{min}, \dots, w\_{max}]$.    
    \item \textbf{Dilation} $d$: Randomly chosen from interval $[1,\dots,  2^{\frac{\log(n-1)}{w-1}}]$. The formula is inherited from the ROCKET-family.
    \item \textbf{Word length} $l$: Randomly chosen from $\{7,8\}$.    
    \item \textbf{Binning strategy}: Randomly chosen from \{"equi-depth", "equi-width"\}.
    \item \textbf{First order differences}: We extract words from both, the raw time series, and its first order difference, effectively doubling the feature space. 
\end{enumerate}

For example, when using $r_{max}=50$ configurations, the feature space has a size of $256 \cdot 50 \approx 10k$, and is roughly in the range of the ROCKET-family. First order differences double the space requirements, effectively generating $20k$ to $70k$ features.
By increasing $r_{max}$ we can increase the size of the feature space. 

We will next introduce the pseudocode of WEASEL 2.0, illustrating the steps outlined before.

\subsection{Pseudocode}\label{sec:pseudocode}

WEASEL 2.0 is based on the dilation mapping, and randomization over different hyper-parameter configurations, where $r_{max}$ controls the number of configurations and thus the number of features generated. 

\begin{algorithm}[!ht]
 	\caption{transform(\\
        D: a list of $m$ data series of length $m$, 
        ${\bf D}=\left({\bf T^{(i)}}\right)_{i \in [1 \dots m]}$;  \\
        $w\_max$ and $w\_min$: the maximal and minimal word length; \\
        $r_{max}$: the ensemble size \\
        )
        } 
 	\label{alg:weasel2.0}
 	\begin{algorithmic}[1]
 		\STATE Let ${\bf H}$ be a vector of word counts of dim $m \times (r_{max} \times 256)$
 		
        \FOR {$r \leftarrow  1$ to $r_{max}$}            
            \STATE $w = random.choose([w\_min,...,w\_max])$
            \STATE $d = random.choose([1, \dots, 2^{\frac{\log(n-1)}{w-1}}])$
            \STATE $l = random.choose([7, 8])$
            \STATE $binning = random.choose(["equidepth", "equiwidth"])$

            \STATE Let ${\bf H'}$ be a vector of word counts with dim $m \times 256$ 

            \FOR {$i \leftarrow  1$ to $m$}                
                \STATE $T' = \textbf{apply\_dilation}(T^{(i)}, d)$
                \FOR {$j \leftarrow 1$ to $n-w+1$}
                   \STATE ${\bf w} \leftarrow$ SFA\_transform($T'_{j:w}, l, \alpha = 2, norm=False$)
 				   \STATE ${H'}^{(i)}[w] \leftarrow {H'}^{(i)}[w] + 1$
                \ENDFOR
            \ENDFOR

            \STATE $H.concatenate(H')$ \COMMENT{ append counts for $r$-th config}        
        \ENDFOR
 	\end{algorithmic}
 \end{algorithm}

Algorithm~\ref{alg:weasel2.0} shows the complete WEASEL 2.0 transform. Given $r_{max}$, $w_{max}$ and $w_{min}$, we randomly choose the parameters for each configuration. Each configuration then generates a dense vector (dictionary) for each of $m$ time series (line 7), with a total size of $m \times 256$. Dilation is applied to each time series through the dilation mapping (line 9). By default, we apply dilation to the first order differences, too, effectively doubling the feature space. Next, windows are extracted from each time series (line 10), and the windows are transformed to words of length $l$ using an alphabet of size $2$ (line 11). Other than in the previous dictionary-base methods using SFA, the alphabet size is fixed to $2$ and the mean-normalization is fixed to $False$ to control the size of the feature space. SFA then chooses the best Fourier coefficients based on maximal variance. I.e. those real and imaginary Fourier values that have the largest spread are favorable to obtain robust words. Next, word counts are increased for each time series (line 12). Finally, the dictionaries of all configurations are concatenated into one dense vector (line 13). 

At the end of transformation, the resulting feature vector is used as input to a RIDGE regression classifier. Its parameters are learned through cross-validation from $alphas=np.logspace(-1, 5, 10)$ and $normalize=False$.

\subsection{Complexity}\label{sec:complexity}

The major advantage of WEASEL 2.0 is its controlled size of the feature space. Depending on the value $r_{max}$, we generate up to $256 \times r_{max}$ distinct words, which can be stored in a dense vector. Using first order differences, we double the number of features to $2 \times 256 \times r_{max}$. For input parameter $r_{max}=50$ we thus generate $25k$ features.
Given $m$ time series, the memory required by WEASEL 2.0 is thus equivalent to $256 \times r_{max} \times m \times 2$ using first order differences. E.g., given a dataset with $m=10k$ and $n=1024$, WEASEL 2.0 needs roughly $976MB$ of memory for $r_{max}=50$. This is comparable to the memory needed by ROCKET et al. (see Section~\ref{sec:feature_space})

The runtime required for classification depends on the machine learner used. Yet, RIDGE regression can be implemented in linear time, depending on the solver used, which makes WEASEL 2.0 also very fast.

\section{Experimental Evaluation}\label{sec:experiments}

\paragraph{Datasets:} We compare our \emph{WEASEL 2.0} classifier to the SotA using the UCR benchmark of $114$ TSC problems~\cite{dau2019ucr}. Each UCR dataset provides a train and test split, which we use unchanged to make our results comparable to prior publications. We visualize comparisons with critical difference diagrams, which compare mean ranks of approaches. A horizontal bar indicates cliques, for which there is no statistical significant difference between approaches in rankings. These cliques are computed using a Wilcoxon-Holm post-hoc analysis and p-value of $0.05$.

\paragraph{Competitors:} We compare WEASEL 2.0 to up to $15$ state-of-the-art TSC methods. \textbf{Dictionary (D):} BOSS, cBOSS, WEASEL, TDE; \textbf{Hybrid (H):} HiveCote 2.0, HiveCote 1.0, TS\_CHIEF, \textbf{Deep-Learning (DL):} InceptionTime; \textbf{Shapelets (S):} R-DST, MrSQM\_SFA\_k5; \textbf{Kernel (K):} Arsenal, MiniRocket, MultiRocket, Rocket, Hydra. We used implementations available in sktime~\cite{loning2019sktime}, or published by the authors~\cite{dempster2022hydra, mrsqm2022, guillaume2022random}. All reported numbers are accuracy on the test split. 

In all figures, we append the names of the methods with its type of approach: D: dictionary, S: Shapelets, K: Kernel, H: Hybrid, DL: Deep Learning.

We will compare methods on (a) accuracy, (b) runtimes, (c) memory-footprint, and (d) the dependency on application domains.

\paragraph{Hardware:}
All experiments ran on a server running LINUX with four 10 core Intel(R) Xeon(R) CPU E7-4870 at 2.40GHz, using sktime v1.4 and python 3.8.3. We also CPU time as runtime of all implementations, to address parallel and single threaded codes. 

To ensure reproducible results, we provide the WEASEL 2.0 source code and the raw measurement sheets.~\cite{WEASELWebPagev2}. WEASEL 2.0 applies the rule of thumb presented in Section~\ref{sec:ensemble} to set the two hyper-parameters $r_{max}$ and $w_{max}$, and all other parameters are set through randomization.

\subsection{Accuracy}

\begin{figure}[t]
\begin{centering}
\includegraphics[width=1\columnwidth]{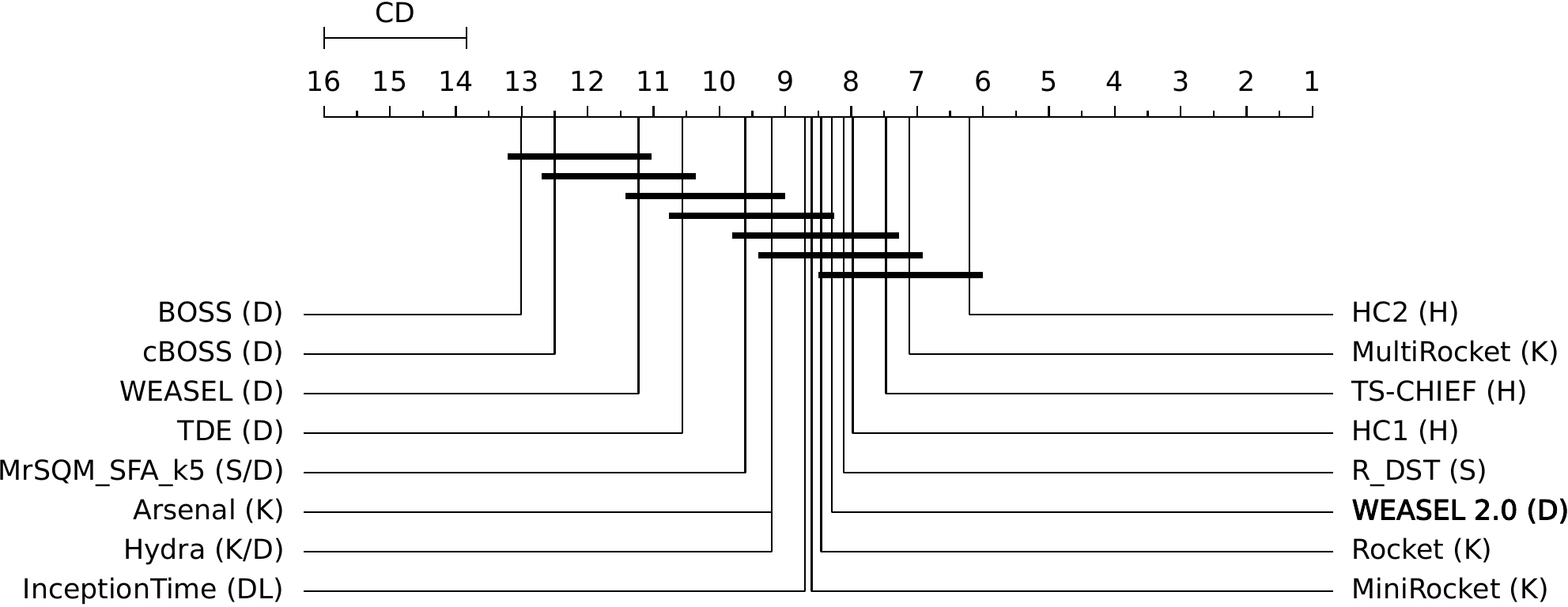}
\end{centering}
\caption{Critical difference plot on average ranks on test accuracy on 114 UCR datasets. Smaller is better. \label{fig:UCR_critical_accuracy}}
\end{figure}

Figure~\ref{fig:UCR_critical_accuracy} shows a critical difference diagram on the average ranking of each competitor method on the 114 datasets. Lower ranks indicate that a method is better, and a horizontal bar indicates that two methods are not  significantly different. WEASEL 2.0 is significantly better than WEASEL, and the most accurate dictionary method (compare Figure~\ref{fig:UCR_critical_accuracy_subset}). WEASEL 2.0 is further in the same group as other non-ensemble dilation-based approaches, such as ROCKET, MiniRocket, R-DST. Among the most accurate methods are hybrids such as HC2, HC1 and TS-CHIEF, which ensembles include variants of dictionary classifiers. Thus, WEASEL 2.0 could be a promising candidate to further improve these.

\begin{figure}[t]
\begin{centering}
\includegraphics[width=1\columnwidth]{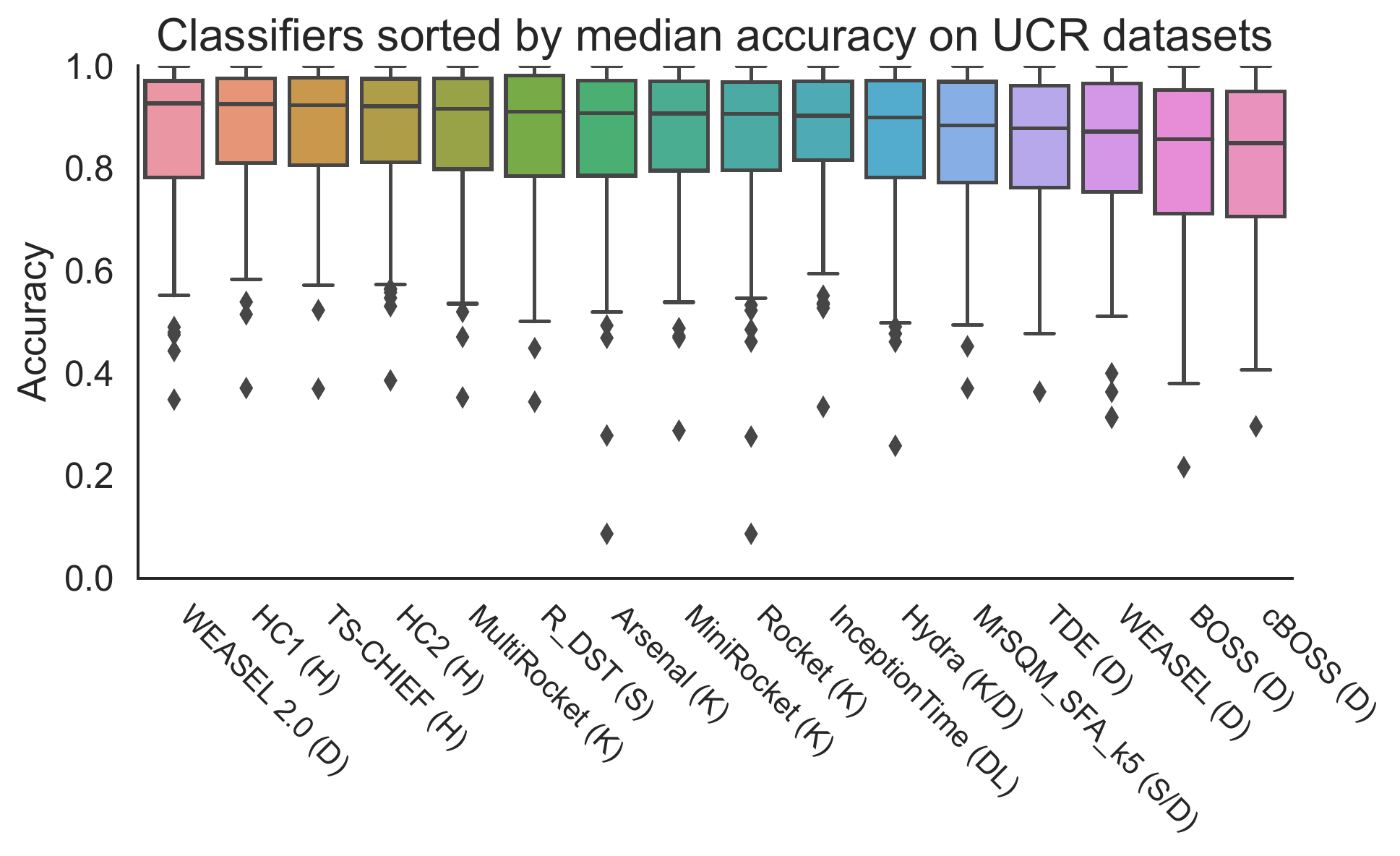}
\end{centering}
\caption{Box-plot on test accuracy on 114 UCR datasets. Methods are sorted by median accuracy.\label{fig:UCR_boxplot_accuracy}}
\end{figure}

The box plots in Figure~\ref{fig:UCR_boxplot_accuracy} show how close the accuracy of current SotA approaches has become. Approaches in this figure are sorted by median accuracy. Top ranking approaches have a low IQR (inter-quartile range), with few outliers, and the majority of datasets scores above $90\%$. 

\begin{figure*}[t]
\begin{centering}
\includegraphics[width=0.6\columnwidth]{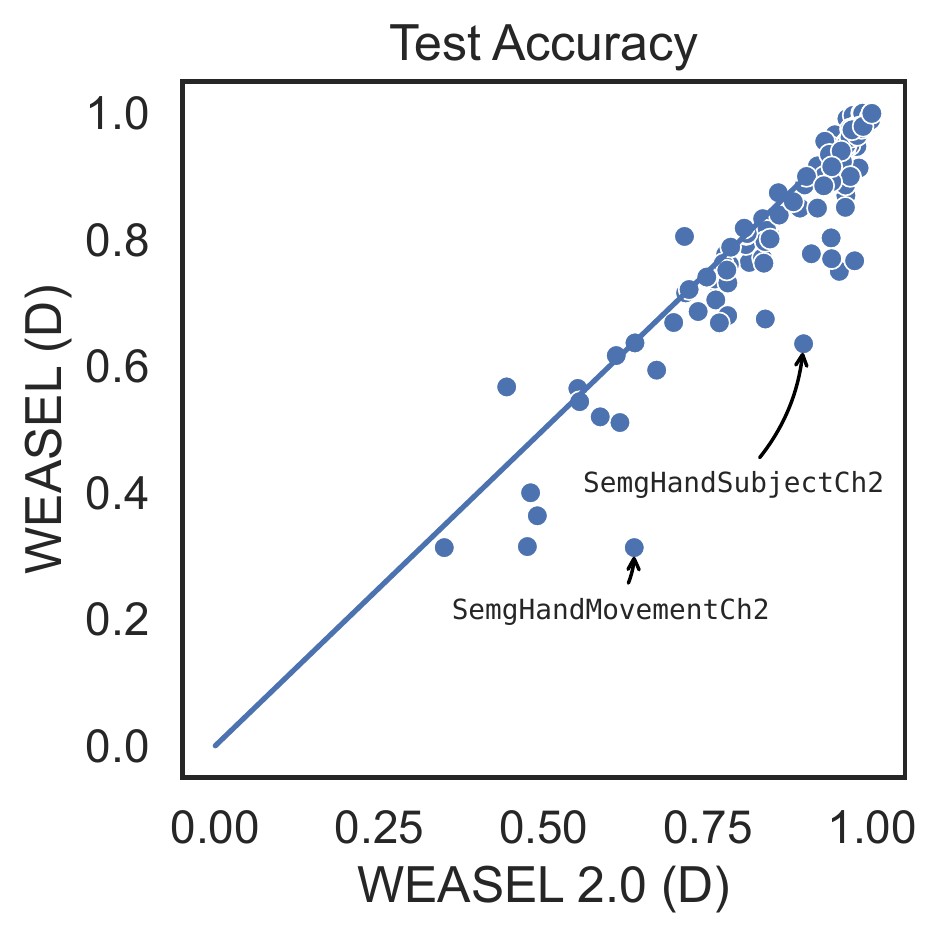}
\includegraphics[width=0.6\columnwidth]{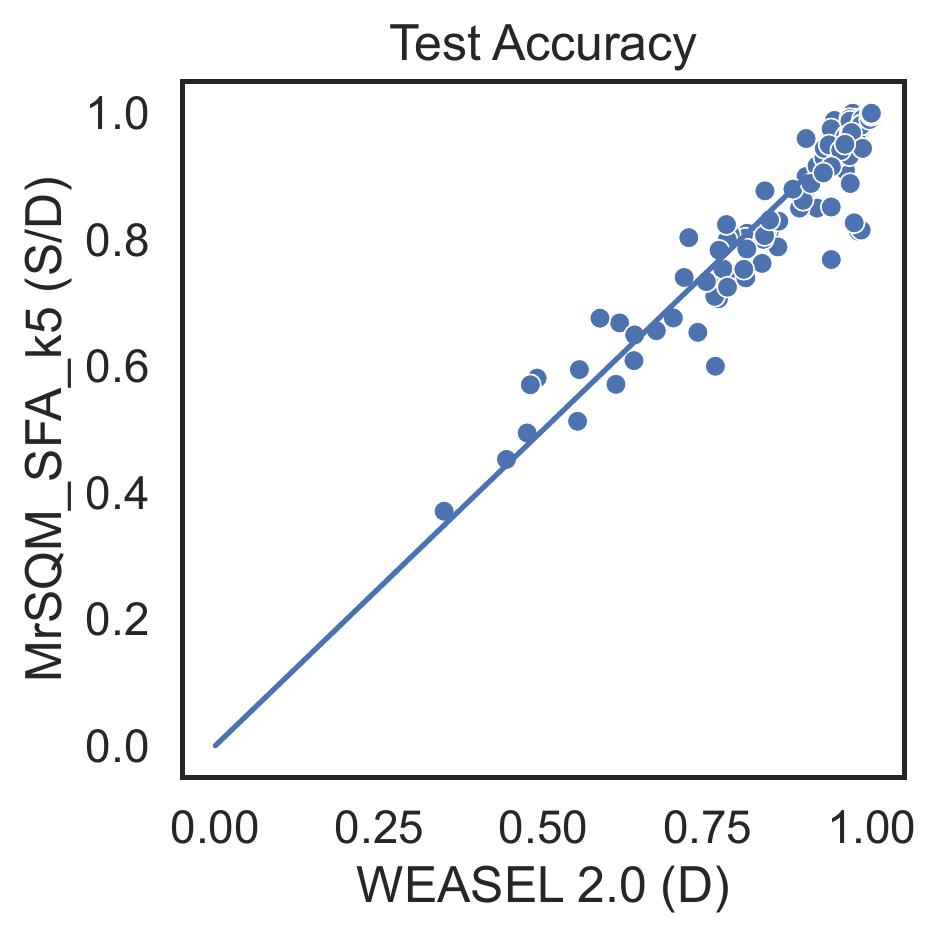}
\includegraphics[width=0.6\columnwidth]{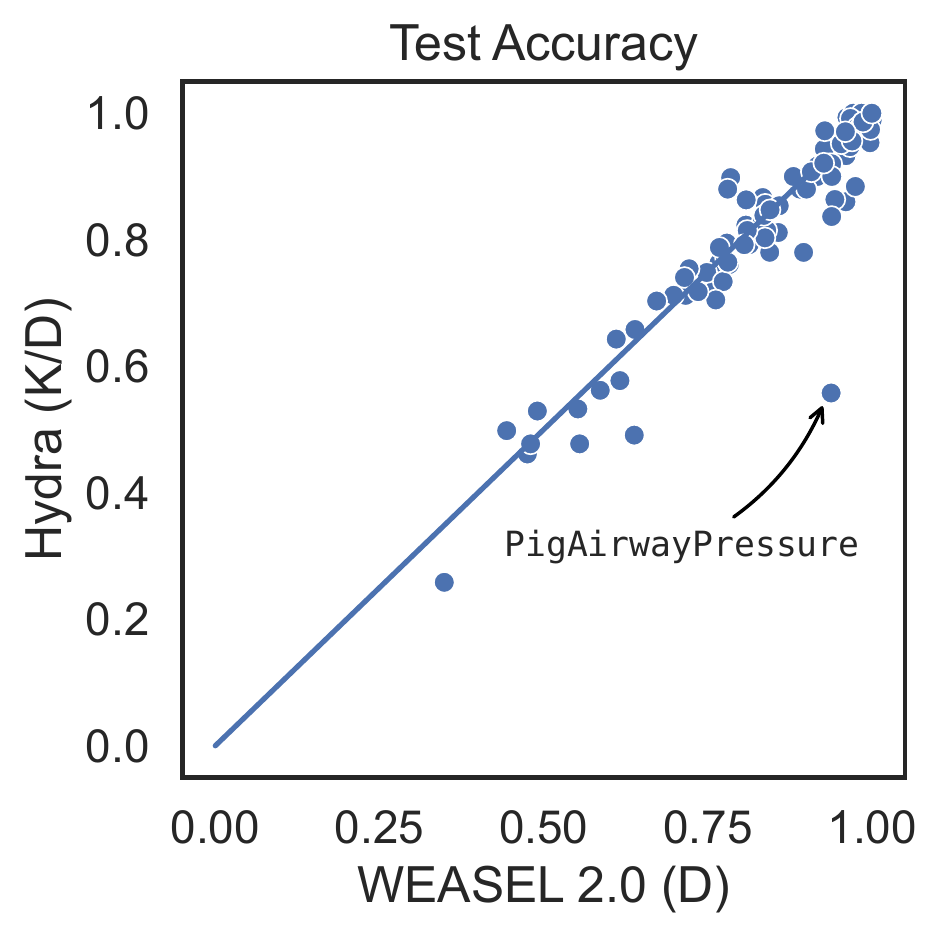}\\
\includegraphics[width=0.6\columnwidth]{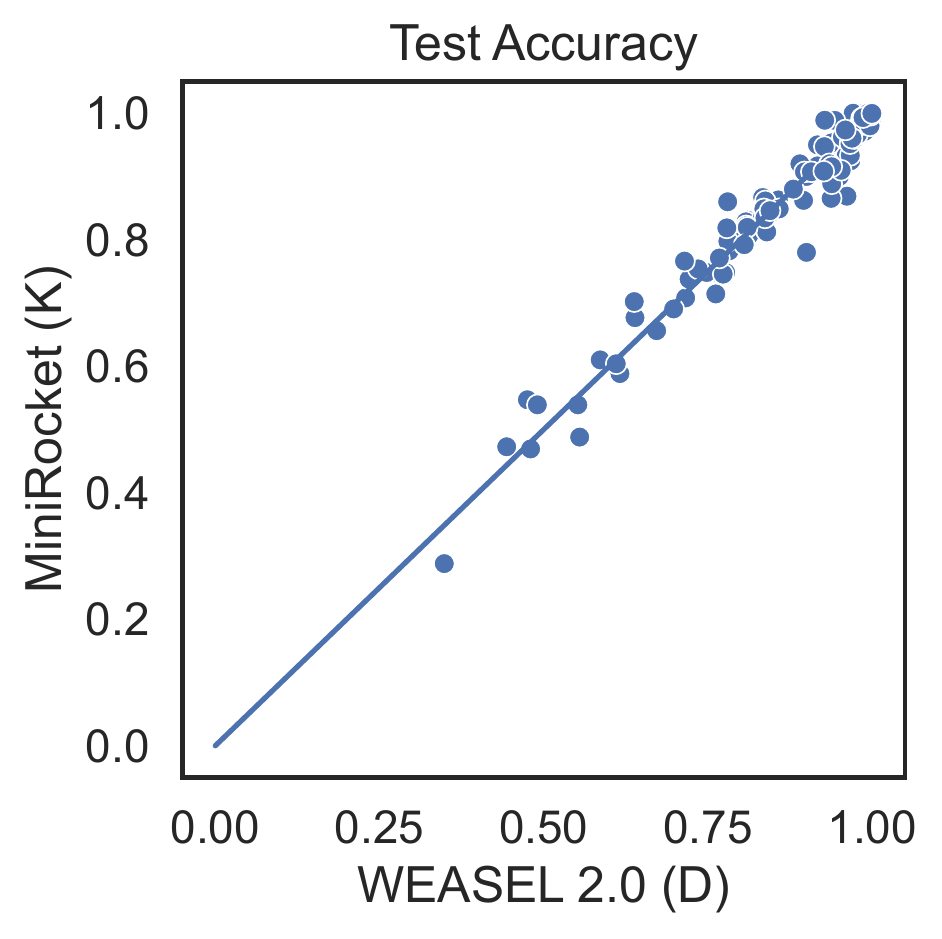}
\includegraphics[width=0.6\columnwidth]{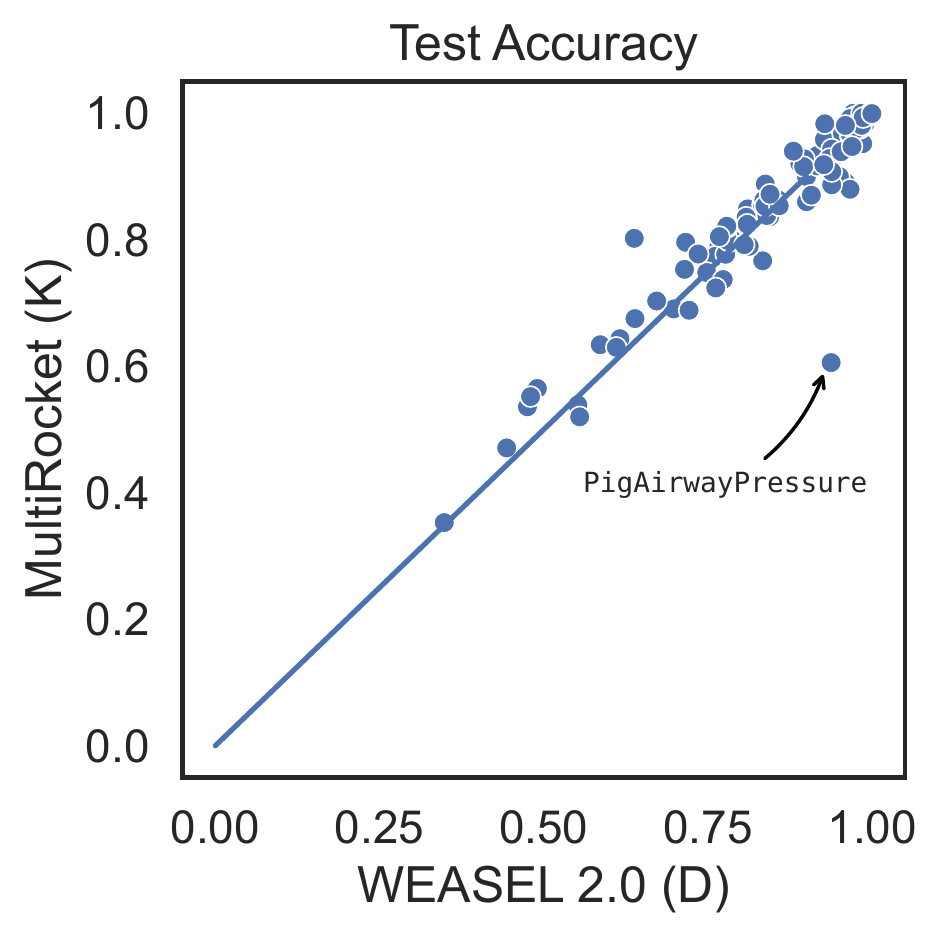}
\includegraphics[width=0.6\columnwidth]{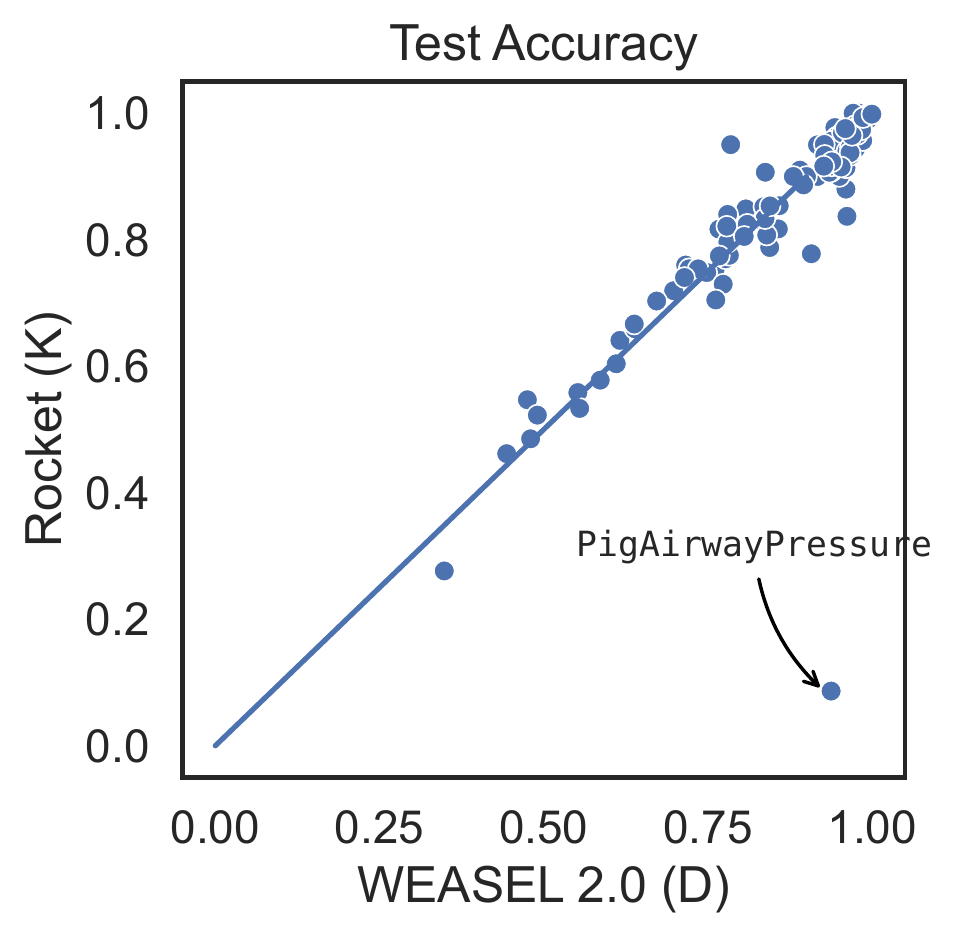}
\end{centering}
\caption{Pairwise comparison of approaches. Each dataset account for one point. A point below (above) the diagonal indicates that WEASEL is more (less) accurate than its competitor on one dataset. \label{fig:UCR_pairwise}}
\end{figure*}

Figure~\ref{fig:UCR_pairwise} shows a pairwise comparison of WEASEL 2.0 to selected competitors. Each point represents the test accuracy of one dataset. A point below (above) the diagonal indicates that WEASEL 2.0 scores higher (lower) than its competitor. Firstly, WEASEL 2.0 performs much better than WEASEL. Most points are below the diagonal. Secondly from the other competitors most points are along or slightly above/below the diagonals. When compared to the Rocket-family the \emph{PigAirwayPressure} datasets stands out. On this dataset, WEASEL 2.0 has a much higher test accuracy than its competitors. It is difficult to identify the property that make ROCKET fail, but this dataset has $52$ classes with tiny differences, which makes it stand out from others.

\subsection{Accuracy by Dataset and by Domain}

\begin{figure*}[t]
\includegraphics[width=2\columnwidth]{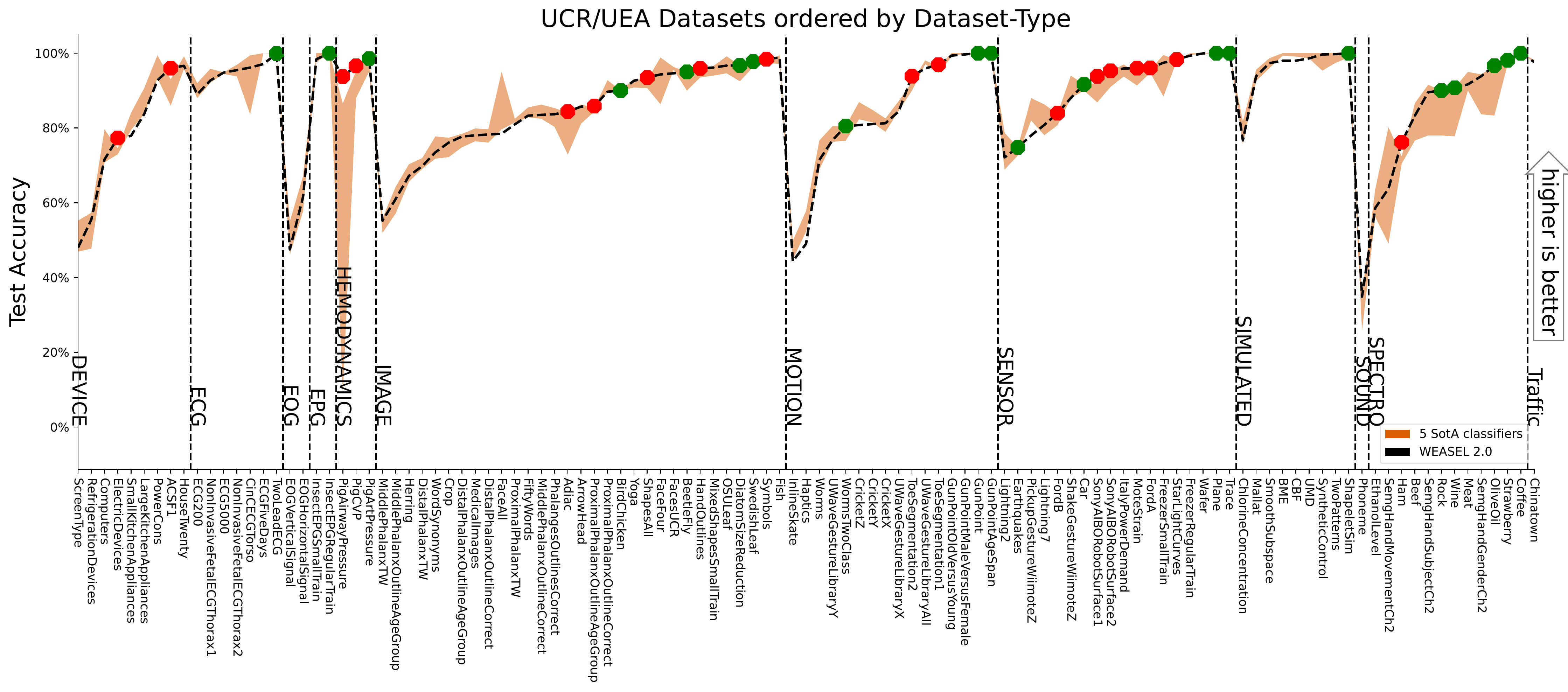}
\caption{Classification accuracies for WEASEL 2.0 vs the best five core classifiers (R-DST, MultiRocket, MiniRocket, Rocket, and Hydra). The orange area represents the six core classifiers' range of accuracies. Red (green) dots indicate where WEASEL 2.0 wins (evens out) against the other classifiers.\label{fig:mine-vs-all}}
\end{figure*}

In this experiment we inspect the performance of WEASEL 2.0 by type of application domain. We grouped datasets by type to test  domain-dependent strengths or weaknesses. We used the predefined grouping of the benchmark data types as defined by~\cite{dau2019ucr}

For this experiment, we only consider the top-$5$ non-ensemble competitors. Figure~\ref{fig:mine-vs-all} shows the accuracies of WEASEL 2.0 (black line) vs. the five top core classifiers R-DST, MultiRocket, MiniRocket, Rocket, and Hydra (orange area). Overall, the performance of WEASEL 2.0 is very competitive for almost all datasets.  
WEASEL 2.0 has the highest percentage of wins in the groups of Hemodynamics, Image Outlines, Motion, Sensor and Spectro. Overall WEASEL 2.0 has $18$ wins (red), and $38$ top scores (red and green). It is only super-seeded by MultiRocket which has $25$ wins and $46$ top scores, see Table~\ref{table:wins_ties}. 

\begin{table}
\small
\begin{tabular}{|c|c|c|c|c|c|c|}
\hline 
 & WEASEL 2.0 & MultiR. & MiniR. & Rocket & Hydra & R-DST\tabularnewline
\hline 
\hline 
win/tie & \textbf{18 / 38} & 25 / 46 & 2 / 22 & 10 / 31 & 5 / 29 & 18 / 38\tabularnewline
\hline 
\end{tabular}\caption{Wins and ties by method on 114 UCR datasets.\label{table:wins_ties}}
\end{table}



\subsection{Scalability}

\begin{figure}[ht]
\begin{centering}
\includegraphics[width=1.0\columnwidth]{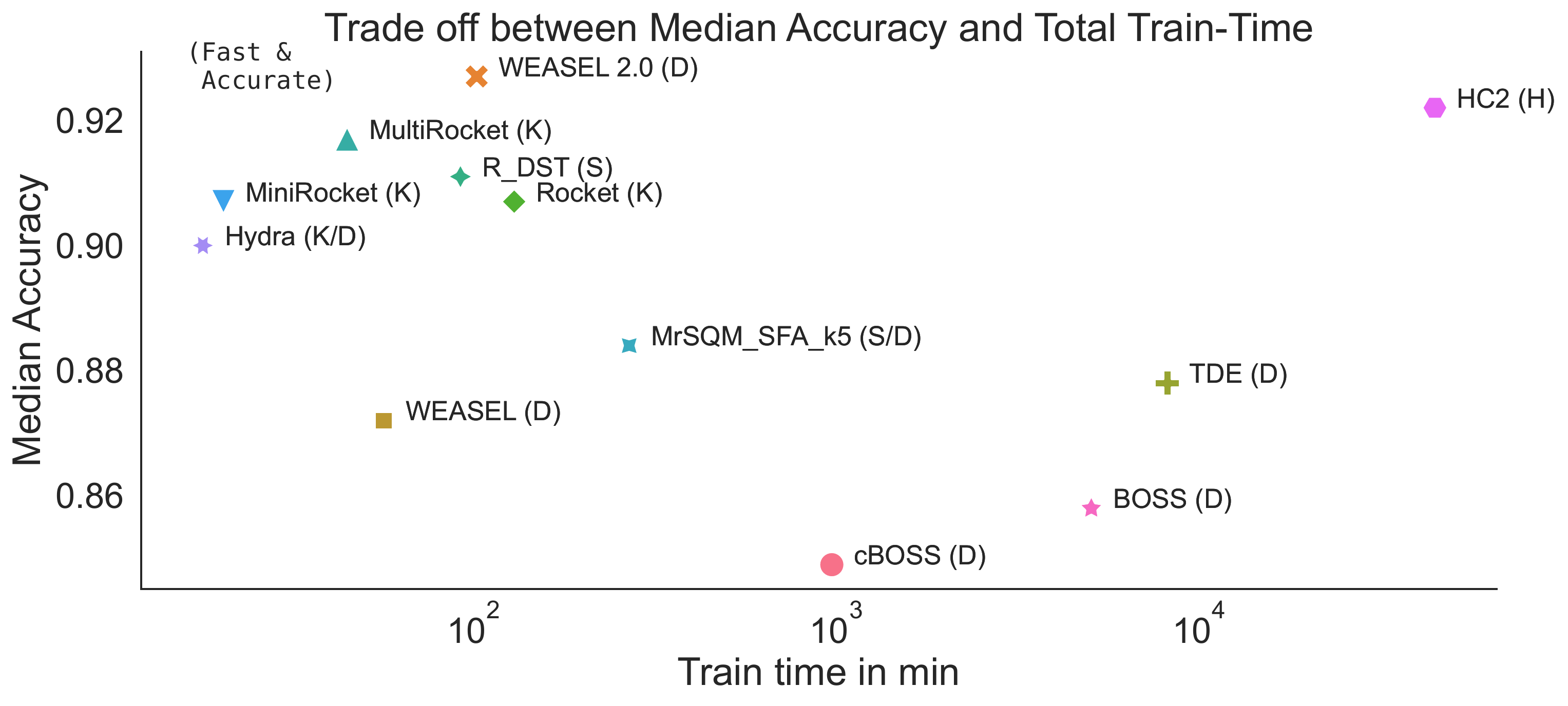}\\
\includegraphics[width=1.0\columnwidth]{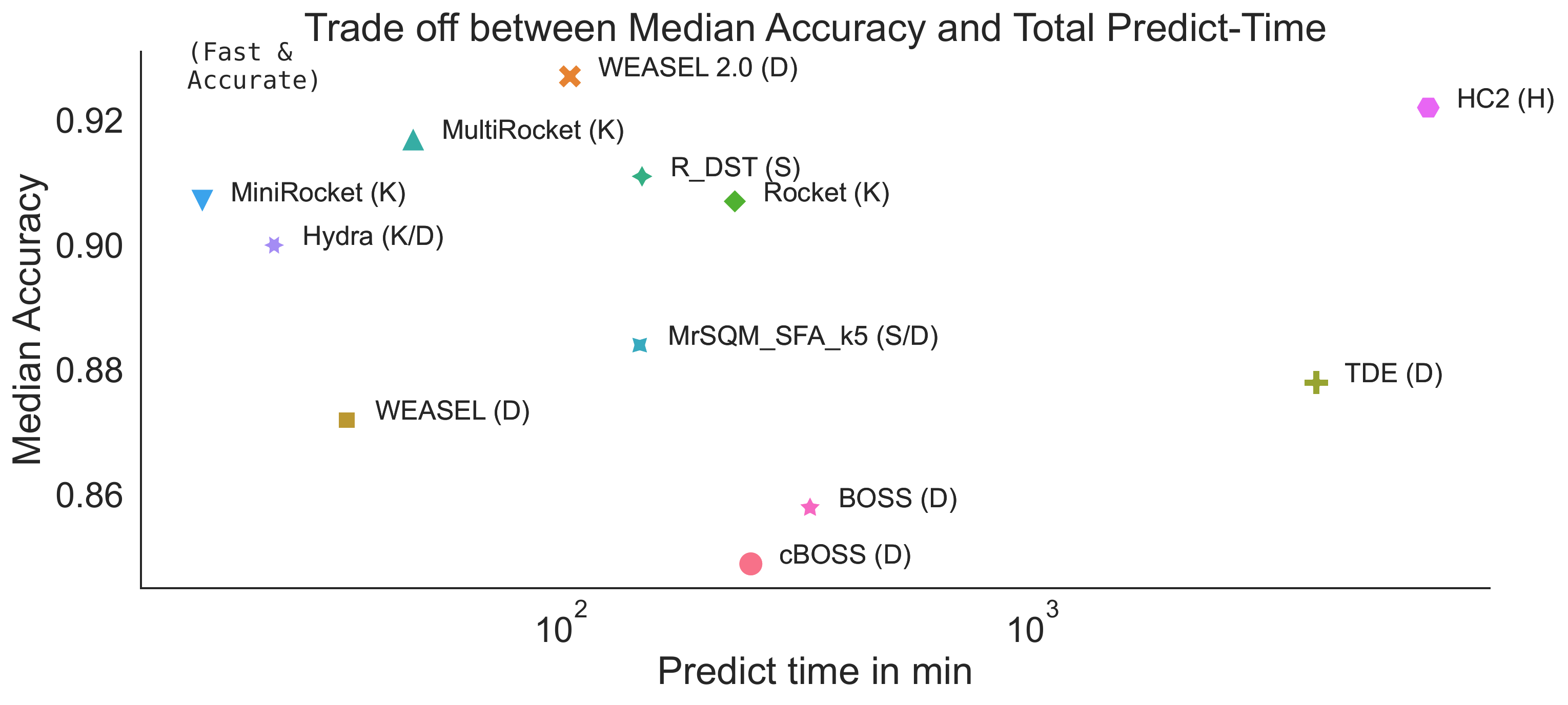}
\end{centering}
\caption{Total runtime on $114$ UCR datasets by approach in minutes against median accuracy.\label{fig:UCR_fit}}
\end{figure}

Figure~\ref{fig:UCR_fit} shows the runtimes for training and predictions of each competitor for non-hybrid methods. Hybrid methods typically have a runtime that is at least $10-100$ times higher. 
The total fit times vary from $30$ minutes (Hydra, MiniRocket) to $130$ (for TDE) and $730$ (for HC2) hours on the full $114$ UCR datasets. WEASEL 1.0 requires a total of $1$ hour, and WEASEL 2.0 requires $1.6$ hours. Predict times are in a similar range of a total of $30$ minutes (MiniRocket), $1.7$ hours for WEASEL 2.0 and $66$ (for TDE) or $110$ (for HC2) hours. Thus, while runtimes have not improved for WEASEL 2.0 over WEASEL 1.0, it has the advantage of a constant sized dictionary, and the SotA accuracy. We will highlight the differences in the memory-footprint in the following Figure~\ref{fig:UCR_memory}.

Figure~\ref{fig:UCR_fit} plots the trade-off between the median accuracy and total runtime (train+predict). The most desirable method has a low runtime and a high accuracy (top left corner). The best Pareto-optimal methods, closest to the upper left edge, are WEASEL 2.0, MultiRocket, MiniRocker and Hydra. Inferior methods are BOSS, TDE or HC 2.0. 

\subsection{Size of Feature Space}\label{sec:feature_space}

\begin{figure}[ht]
\begin{centering}
\includegraphics[width=1.0\columnwidth]{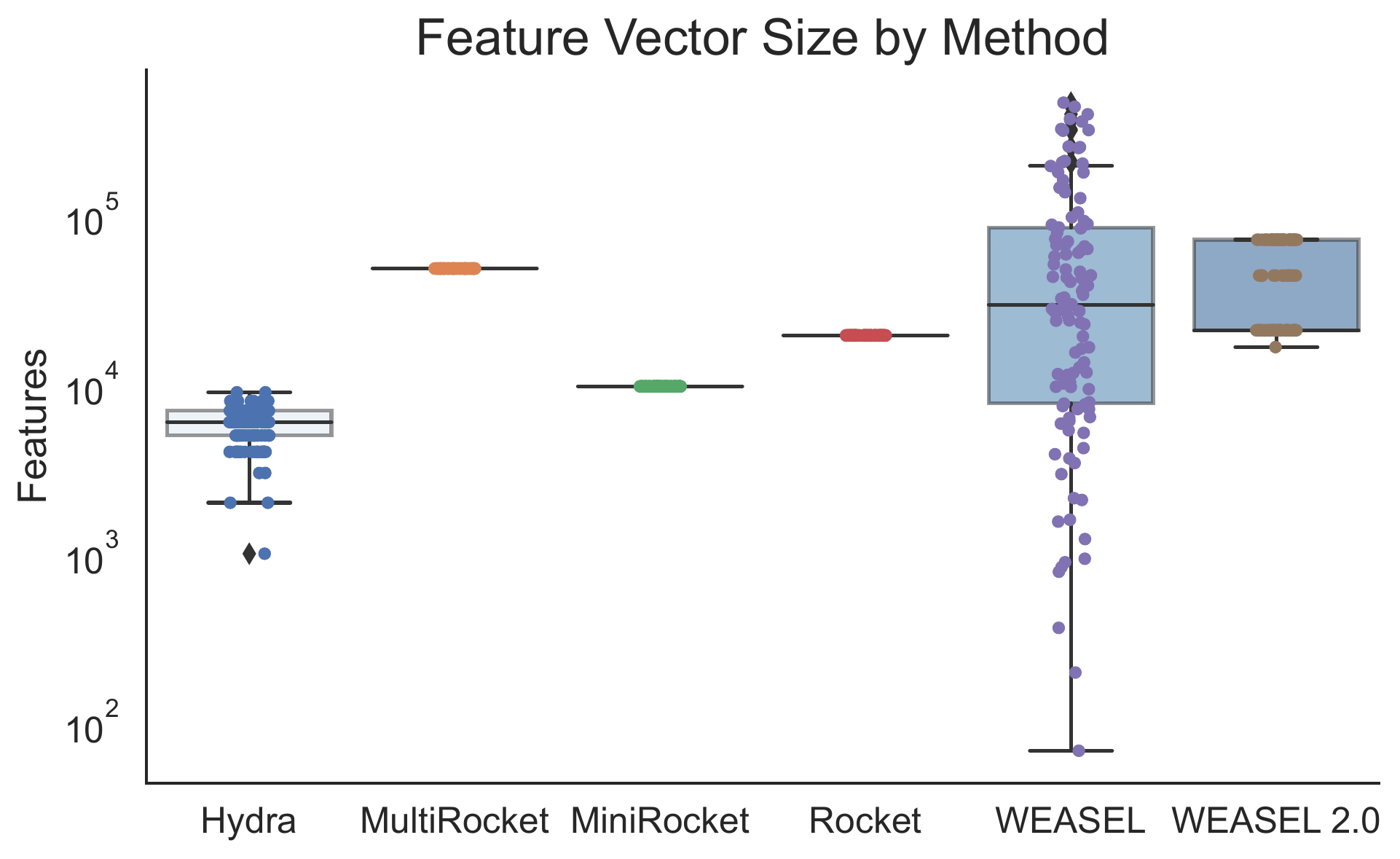}
\end{centering}
\caption{Size of the feature vector on the UCR datasets. The size of WEASEL 2.0 depends on the parameterization of $r_{max}$, chosen from $3$ values.\label{fig:UCR_memory}}
\end{figure}

Figure~\ref{fig:UCR_memory} shows the size of the feature space per time series compared to the ROCKET family. For WEASEL the size is defined by the size of the dictionary. WEASEL 1.0 showed excessive memory consumption, which made it unusable for many scenarios. Meanwhile, WEASEL 2.0 has a controlled size of the dictionary, which depends on the value of $r_{max}$ and the rule of thumb used (Section~\ref{sec:ensemble}). Depending of the parameterization of $r_{max}=50 (150)$ WEASEL 2.0 generates $25k (76k)$ features.
The size of the feature space for ROCKET and its variants, using default parameters, varies between approaches from $1k$ to $50k$.

\section{Conclusion}\label{sec:conclusion}

In this work, we have presented WEASEL 2.0, a novel TSC method following the dictionary approach, which achieves highly competitive classification accuracy. 
It is fast, and other than its predecessors (BOSS, WEASEL, TDE), has a predictable, constant memory footprint. This makes it applicable in domains with high runtime and accuracy constraints. 

The novelty of WEASEL 2.0 lies in combining dilation and randomization with the dictionary model, along with a carefully refined symbolic representation for extracting words. WEASEL 2.0 uses small, $256$ word, dictionaries derived for each set of input features. While this causes high bias in combination with linear classification, we reduce bias and add variance through ensembling random hyper-parameter configurations. 

In our evaluation on the UCR datasets, WEASEL 2.0 is significantly more accurate than its predecessors and the best in its dictionary-based class. It is in the group of the best and fastest SotA methods, including the ROCKET-family, and has a predictable constant memory footprint.


\bibliographystyle{ACM-Reference-Format}
\balance 
\bibliography{weasel2}

\end{document}